\documentclass[conference]{IEEEtran}

\usepackage[acronym]{glossaries}
\usepackage{cite}
\usepackage{amsmath,amssymb,amsfonts}
\usepackage{graphicx}
\usepackage{xcolor}
\usepackage{booktabs}
\usepackage{array}
\usepackage{microtype}
\usepackage{tikz}
\usepackage{pgfplots}
\usepackage{pgfplotstable}
\usepackage{placeins}
\usepackage{multirow}
\usepackage{balance}

\newacronym{tsfm}{TSFM}{time series foundation model}
\newacronym{peft}{PEFT}{parameter-efficient fine-tuning}
\newacronym{lora}{LoRA}{low-rank adaptation}
\newacronym{qlora}{QLoRA}{quantized low-rank adaptation}
\newacronym{dora}{DoRA}{weight-decomposed low-rank adaptation}
\newacronym{nmae}{NMAE}{normalized mean absolute error}
\newacronym{nrmse}{NRMSE}{normalized root mean square error}
\newacronym{id}{ID}{in-distribution}
\newacronym{ood}{OOD}{out-of-distribution}
\newacronym{gpu}{GPU}{graphics processing unit}

\pgfplotsset{compat=1.18}
\usepgfplotslibrary{groupplots,statistics,fillbetween}
\usetikzlibrary{
    positioning,
    arrows.meta,
    calc,
    fit,
    backgrounds,
    patterns,
    patterns.meta,
    matrix,
    decorations.pathreplacing
}

\graphicspath{{figures/}{figures/results/}}

\newcommand{\nmae}{\operatorname{NMAE}}
\newcolumntype{L}[1]{>{\raggedright\arraybackslash}p{#1}}
\newcolumntype{C}[1]{>{\centering\arraybackslash}p{#1}}

\newcommand{\papergraphic}[3]{%
  \begin{minipage}{\linewidth}
  \centering
  \IfFileExists{#1.tex}{%
    \input{#1}%
  }{\IfFileExists{#1.pdf}{%
    \includegraphics[width=\linewidth,height=#3,keepaspectratio]{#1.pdf}%
  }{\IfFileExists{#1.png}{%
    \includegraphics[width=\linewidth,height=#3,keepaspectratio]{#1.png}%
  }{%
    \fbox{\parbox[c][#3][c]{0.94\linewidth}{%
      \centering\scriptsize
      Missing figure: \texttt{\detokenize{#2}}%
    }}%
  }}}%
  \end{minipage}%
}

\title{Parameter-Efficient Adaptation of Time Series Foundation Models for Electrical Load Forecasting}

\author{
\IEEEauthorblockN{
Aliullah\textsuperscript{1,2}\textsuperscript{*},
Julian Oelhaf\textsuperscript{1}\textsuperscript{*},
Edouard Fouch\'e\textsuperscript{2},
Jessica Deuschel\textsuperscript{2},
Christian Bergler\textsuperscript{3},\\
Andreas Maier\textsuperscript{1},
Siming Bayer\textsuperscript{1}
}
\IEEEauthorblockA{
\textsuperscript{1}Pattern Recognition Lab,
Friedrich-Alexander-Universit\"at Erlangen-N\"urnberg\\
\textsuperscript{2}Siemens AG, Smart Infrastructure\\
\textsuperscript{3}Department of Electrical Engineering, Media and Computer Science,
Ostbayerische Technische Hochschule Amberg-Weiden\\
\textsuperscript{*}Corresponding authors: ali.ullah@fau.de, julian.oelhaf@fau.de
}
}

\begin{document}
\maketitle

\begin{abstract}
Accurate electrical load forecasting supports power-system operation across countries, regions, cities, and buildings, but repeated development of target-specific models is costly. \Glspl{tsfm} provide reusable pretrained backbones, although multi-domain pretraining may not fully capture load-specific characteristics. This paper compares pretrained inference, full fine-tuning, and \gls{lora} in terms of forecasting accuracy and adaptation efficiency across four electrical load aggregation levels. MOIRAI 1.0 Large serves as the primary backbone, while Chronos-2 assesses whether the findings transfer across model families. The evaluation covers both \gls{id} and \gls{ood} settings and examines whether compact adaptation provides consistent gains over strong pretrained checkpoints. Pretrained foundation models outperform the implemented seasonal naive and XGBoost baselines across the evaluated aggregation levels. On MOIRAI Large, rank-16 \gls{lora} reduces \gls{nmae} by 7.5\,\% on average while updating 3.146 million parameters, about 1\,\% of the model, and retains most of the improvement from full fine-tuning. Chronos-2 achieves lower pretrained \gls{nmae} than MOIRAI, while its relative adaptation gains remain smaller overall. \Gls{ood} results show that adaptation benefits vary by target, with the clearest gains for the local-area series and several buildings. Overall, \gls{lora} provides a favorable accuracy--efficiency trade-off when adaptation yields a meaningful gain over a strong pretrained checkpoint.
\end{abstract}


\glsresetall

\section{Introduction}
Accurate electrical load forecasting is essential for reliable and economically viable power-system operation. Short-term forecasts inform generation scheduling, balancing and reserve decisions, market participation, demand response, and the operation of local energy systems and buildings. Forecasting difficulty also changes with aggregation: national and regional demand is comparatively smooth, while city and building loads are more strongly affected by weather, occupancy, operating schedules, and local equipment behavior \cite{sevlian2018,peng2019}.

In practice, many systems, market areas, customer groups, and buildings require individual load forecasts. Conventional forecasting approaches can provide strong local performance, but they often require models to be trained separately for individual series or datasets \cite{arango2025chronosx}. As the number of forecasting targets grows, model configuration and training must therefore be repeated across more series. For comparison, a weekly seasonal naive model provides a meaningful reference because electrical load exhibits daily, weekly, and seasonal structure, with many series showing strong weekly repetition. Gradient-boosted trees provide a supervised machine-learning alternative using lagged load and external covariates \cite{chen2016xgboost}. Both approaches can be effective, but target-specific models and feature pipelines limit their scalability across large forecasting collections.
 
\Glspl{tsfm} offer a reusable pretrained alternative. \glspl{tsfm} aim to transfer temporal representations learned from large and heterogeneous datasets. MOIRAI, a \gls{tsfm}, provides a universal and covariate-aware architecture for forecasting across frequencies and dimensions \cite{woo2024moirai}. Chronos formulates forecasting as token prediction \cite{ansari2024chronos}, and Chronos-2 extends this model family to univariate, multivariate, and covariate-informed forecasting \cite{ansari2025chronos2}. Recent peak-aware evaluation across area-code, secondary-substation, and low-voltage-feeder levels found that Chronos-2 performed strongly during high-demand periods across all three distribution-grid aggregation levels \cite{chattopadhyay2026peak}. A strong pretrained checkpoint can therefore provide zero-shot forecasts, meaning forecasts without target-specific training, across diverse electrical load series.

The remaining question is how much downstream adaptation such a pretrained model requires for electrical load forecasting. Three practical options exist: (i) pretrained inference applies the pretrained checkpoint without downstream training; (ii) full fine-tuning specializes the complete model by updating all model parameters; and (iii) \gls{lora} provides an intermediate option by freezing the pretrained backbone and learning compact low-rank updates in selected transformations \cite{hu2022lora}. \Gls{qlora} combines low-rank adaptation with a quantized backbone \cite{dettmers2023qlora}, and \gls{dora} decomposes weight updates into directional and magnitude components \cite{liu2024dora}. The present work focuses on \gls{lora} because the primary question is not which \gls{peft} variant performs best, but whether compact adaptation retains sufficient forecasting accuracy to provide an operationally attractive alternative to full fine-tuning. The value of \gls{lora} therefore depends on whether it retains the relevant accuracy gains of full fine-tuning while updating far fewer task-specific parameters.

Existing work shows that pretrained models can be competitive in electricity and building load forecasting and that lightweight adaptation can be effective in narrower settings \cite{gupta2024lora,park2025building}. However, it remains unclear when \gls{peft} is preferable to pretrained inference or full fine-tuning across electrical load aggregation levels, and whether adaptation gains transfer to unseen systems and buildings. This study addresses that gap through a systematic benchmark of pretrained inference, full fine-tuning, and \gls{lora} across country-level, regional-level, city-level, and building-level load series. MOIRAI 1.0 Large serves as the primary backbone, while Chronos-2 tests whether the central findings transfer to a second model family. The benchmark combines \gls{id} and \gls{ood} evaluation with analyses of adapter rank, \gls{peft} method, training-data composition, trainable parameters, and \gls{gpu} configurations.

\section{Methodology}
This section describes the datasets, forecasting setup, model configurations, adaptation methods, and evaluation protocols across four electrical load aggregation levels.

\subsection{Data and Forecasting Setup}
The study compiles the experimental data from three sources. The Worldwide Electricity Load Dataset provides the country-level, regional-level, and most city-level series \cite{ullah2026weld}. The collection is supplemented with three city series from the Power Consumption of Tetouan City dataset \cite{salam2018tetouan} and building series from Building Data Genome 2 \cite{miller2020bdg2}.  
Depending on the source, the available observations cover periods between 2016 and 2025.

The preprocessing pipeline aligns all sources to a common hourly schema. It excludes series with more than 20\,\% missing observations or runs of at least 24 identical values. Observations with an absolute series-level z-score above six are treated as missing, and the remaining gaps are filled using preceding values only. After these quality checks, the main collection contains 10 national systems, 13 regional or system-level series, 22 local electricity market zones, the three Tetouan zones, and 36 educational and office buildings. Temperature over the forecast horizon serves as a future covariate in the city-level and building-level experiments.

The training setup divides each series chronologically into approximately 80\,\% training, 10\,\% validation, and 10\,\% test data. Forecast targets remain within each partition, while historical context may extend into preceding partitions. The experiments use a context length of 2048 hours, matching the initial Chronos-2 pretraining context \cite{ansari2025chronos2}, together with a 96-hour forecast horizon and a stride of 96.

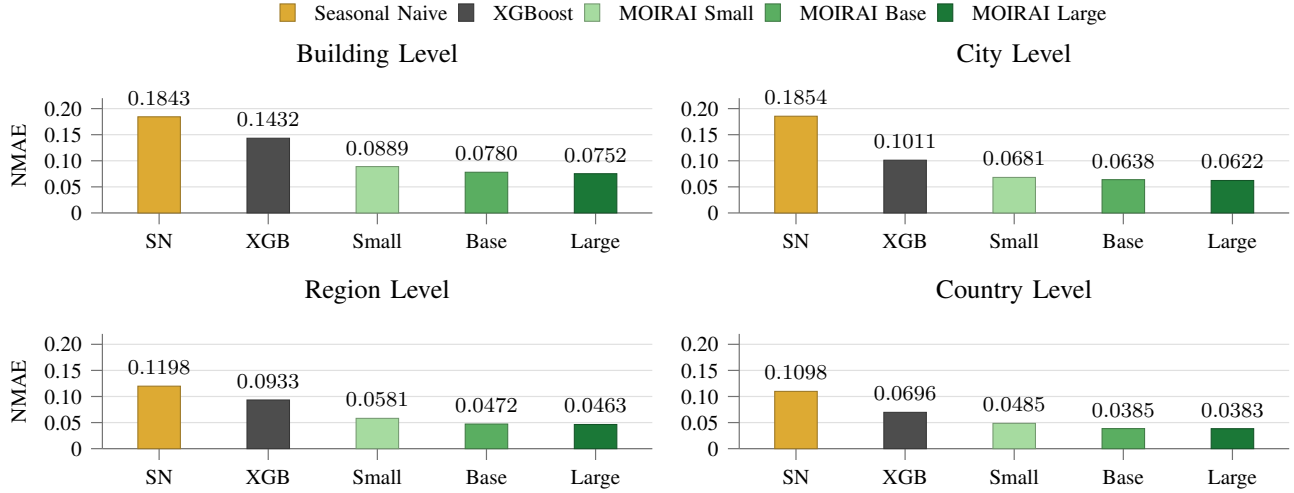
\begin{figure*}[t]
\centering
\centering

\definecolor{seasonal}{HTML}{DDAA33}
\definecolor{xgboost}{HTML}{4D4D4D}
\definecolor{smallmodel}{HTML}{A6DBA0}
\definecolor{basemodel}{HTML}{5AAE61}
\definecolor{largemodel}{HTML}{1B7837}

\newcommand{\plotmoiraimethods}[5]{%
    \addplot+[
        fill=seasonal,
        draw=seasonal!70!black
    ] coordinates {(SN,#1)};

    \addplot+[
        fill=xgboost,
        draw=xgboost!70!black
    ] coordinates {(XGB,#2)};

    \addplot+[
        fill=smallmodel,
        draw=smallmodel!70!black
    ] coordinates {(Small,#3)};

    \addplot+[
        fill=basemodel,
        draw=basemodel!70!black
    ] coordinates {(Base,#4)};

    \addplot+[
        fill=largemodel,
        draw=largemodel!70!black
    ] coordinates {(Large,#5)};
}

\begin{tikzpicture}
    \begin{groupplot}[
        group style={
            group size=2 by 2,
            horizontal sep=1.15cm,
            vertical sep=1.6cm,
        },
        width=\columnwidth,
        height=0.35\columnwidth,
        ybar=0pt,
        /pgf/bar width=16pt,
        /pgf/bar shift=0pt,
        ymin=0,
        ymax=0.22,
        ytick={0,0.05,0.10,0.15,0.20},
        yticklabels={0,0.05,0.10,0.15,0.20},
        yticklabel style={
            font=\footnotesize
        },
        scaled y ticks=false,
        symbolic x coords={
            SN,
            XGB,
            Small,
            Base,
            Large
        },
        xtick={
            SN,
            XGB,
            Small,
            Base,
            Large
        },
        xticklabels={
            SN,
            XGB,
            Small,
            Base,
            Large
        },
        x tick label style={
            font=\footnotesize
        },
        tick align=outside,
        axis x line*=bottom,
        axis y line*=left,
        axis line style={
            draw=black!55,
            line width=0.45pt
        },
        tick style={
            draw=black!55,
            line width=0.45pt
        },
        ymajorgrids=true,
        grid style={
            draw=black!12,
            line width=0.45pt
        },
        enlarge x limits=0.13,
        nodes near coords={%
            \pgfmathprintnumber[
                fixed,
                precision=4,
                fixed zerofill
            ]{\pgfplotspointmeta}%
        },
        nodes near coords style={
            font=\footnotesize,
            text=black,
            yshift=1pt
        },
        title style={
            font=\normalfont,
            yshift=2pt
        },
        ylabel style={
            font=\footnotesize
        },
        clip=false,
    ]

    \nextgroupplot[
        title={Building Level},
        ylabel={NMAE},
        legend to name=moiraicommonlegend,
        legend columns=5,
        legend style={
            draw=none,
            font=\footnotesize,
            column sep=0.18cm,
            /tikz/every even column/.append style={
                column sep=0.07cm
            },
            legend image code/.code={%
                \draw[##1]
                (0,-0.09cm)
                rectangle
                (0.20cm,0.14cm);%
            },
        },
    ]

    \plotmoiraimethods
        {0.1843}
        {0.1432}
        {0.0889}
        {0.0780}
        {0.0752}

    \addlegendentry{Seasonal Naive}
    \addlegendentry{XGBoost}
    \addlegendentry{MOIRAI Small}
    \addlegendentry{MOIRAI Base}
    \addlegendentry{MOIRAI Large}

    \nextgroupplot[
        title={City Level},
    ]

    \plotmoiraimethods
        {0.1854}
        {0.1011}
        {0.0681}
        {0.0638}
        {0.0622}

    \nextgroupplot[
        title={Region Level},
        ylabel={NMAE},
    ]

    \plotmoiraimethods
        {0.1198}
        {0.0933}
        {0.0581}
        {0.0472}
        {0.0463}

    \nextgroupplot[
        title={Country Level}
    ]

    \plotmoiraimethods
        {0.1098}
        {0.0696}
        {0.0485}
        {0.0385}
        {0.0383}

    \end{groupplot}

    \node[anchor=south] at
    ($(group c1r1.north east)!0.5!
      (group c2r1.north west)+(0,0.72cm)$)
    {\pgfplotslegendfromname{moiraicommonlegend}};

\end{tikzpicture}
\caption{Mean \gls{nmae} of the weekly seasonal naive baseline, XGBoost, and pretrained MOIRAI variants across the four electrical load aggregation levels. Lower values indicate better forecasting performance.}
\label{fig:baseline}
\end{figure*}

\subsection{Models and Adaptation}

The weekly Seasonal Naive baseline predicts each target from the observation recorded 168 hours earlier. The gradient-boosted tree baseline, implemented with XGBoost, uses load lags of 1, 24, 168, and 336 hours, together with temperature for the city-level and building-level series, and generates the 96-hour forecast recursively. The pretrained model pool includes MOIRAI 1.0 Small, Base, and Large. The study retains MOIRAI 1.0 Large, which contains approximately 311 million parameters, for systematic adaptation because it achieves the strongest pretrained result within the MOIRAI family. Cross-model replication uses the publicly released Chronos-2 model, which contains approximately 120 million parameters.

Full fine-tuning updates all trainable model parameters. For MOIRAI, the \gls{peft} configuration inserts \gls{lora} into the query, key, value, and output projections of the attention layers. Rank 16 serves as the primary configuration because it is the smallest evaluated rank and updates approximately 3.146 million parameters, corresponding to about 1\,\% of MOIRAI Large. Ranks 32, 64, and 128 are evaluated as supporting ablations to determine whether additional adapter capacity improves forecasting performance.

The MOIRAI experiments use AdamW, validation-loss checkpoint selection, and early stopping. Full fine-tuning follows the official Uni2TS configuration released with MOIRAI and uses a learning rate of $10^{-5}$ \cite{woo2024moirai}. For \gls{lora}, a learning rate of $10^{-4}$ is used as a conservative task-specific configuration informed by the original \gls{lora} methodology \cite{hu2022lora}. The Chronos-2 full fine-tuning and \gls{lora} experiments use 2000 optimization steps, FP16 precision, and the same context length and forecast horizon. Supporting ablations additionally compare \gls{lora}, \gls{qlora}, and \gls{dora}, examine building-level training coverage, and evaluate the same \gls{lora} ranks for Chronos-2.

\subsection{Evaluation}
The \gls{id} evaluation uses chronological test observations from series that contribute training windows during adaptation. Test observations are excluded from optimization, early stopping, and checkpoint selection. The Chronos-2 rank comparison is retrospective and reports the lowest test \gls{nmae} at each aggregation level. The evaluation datasets are not explicitly listed in Chronos-2's reported pretraining data, while MOIRAI includes BDG2 subsets, so pretraining overlap cannot be ruled out. The \gls{ood} evaluation comprises eight series excluded from downstream adaptation and checkpoint selection: one national system, one regional system, one local operational area, two healthcare buildings, and three residential buildings. The pretrained, \gls{lora}-adapted, and fully fine-tuned checkpoints are applied without additional training using the same fixed 2048-hour context as the \gls{id} experiments. All four \gls{lora} ranks were evaluated retrospectively on each \gls{ood} series, with the lowest observed error reported for each target and metric.

The evaluation uses \gls{nmae} as the primary metric. The \gls{ood} analysis additionally reports \gls{nrmse}, which gives greater weight to larger forecast errors. For each series $k$,
\begin{equation}
\nmae_k=
\frac{\frac{1}{n_k}\sum_{i=1}^{n_k}|y_{k,i}-\hat{y}_{k,i}|}
{\max_{1\leq i\leq n_k}|y_{k,i}|},
\label{eq:nmae}
\end{equation}
where $n_k$ denotes the number of evaluated observations, $y_{k,i}$ denotes the observed load, and $\hat{y}_{k,i}$ denotes the corresponding point forecast. Equation~\eqref{eq:nmae} defines the error for each test series. The aggregation procedure then averages the resulting series-level \gls{nmae} values within each aggregation level, giving every series equal weight. Probabilistic models contribute their median forecasts as point predictions.

Parameter efficiency is reported using the number of trainable parameters and their percentage of the model's total parameter count. The \gls{gpu} configuration used for each adaptation method is reported for context.

\section{Results and Discussion}
The results follow five stages: pretrained backbone selection, comparison of full fine-tuning and \gls{lora} for MOIRAI, replication of the central comparison with Chronos-2, \gls{ood} transfer, and derivation of a practical model-selection rule.

\subsection{Pretrained Performance and Backbone Selection}
The first stage evaluates whether pretrained \glspl{tsfm} provide useful electrical load forecasts before downstream adaptation. Fig.~\ref{fig:baseline} compares Seasonal Naive, XGBoost, and the three pretrained MOIRAI variants across the four aggregation levels. Every pretrained MOIRAI configuration achieves lower \gls{nmae} than both implemented baselines. This result shows that broad pretraining transfers effectively to the evaluated electrical load series and provides a meaningful starting point before adaptation is considered.

MOIRAI Large gives the strongest pretrained result within its family, with mean \gls{nmae} values of approximately 0.0752, 0.0622, 0.0463, and 0.0383 for building-level, city-level, regional-level, and country-level data, respectively. The results also follow the expected aggregation pattern: country-level and regional-level load series are easier to forecast than local building load series because aggregation smooths individual operating events. This result motivates the selection of MOIRAI Large for the primary adaptation study.

\subsection{MOIRAI Adaptation Accuracy and Efficiency}
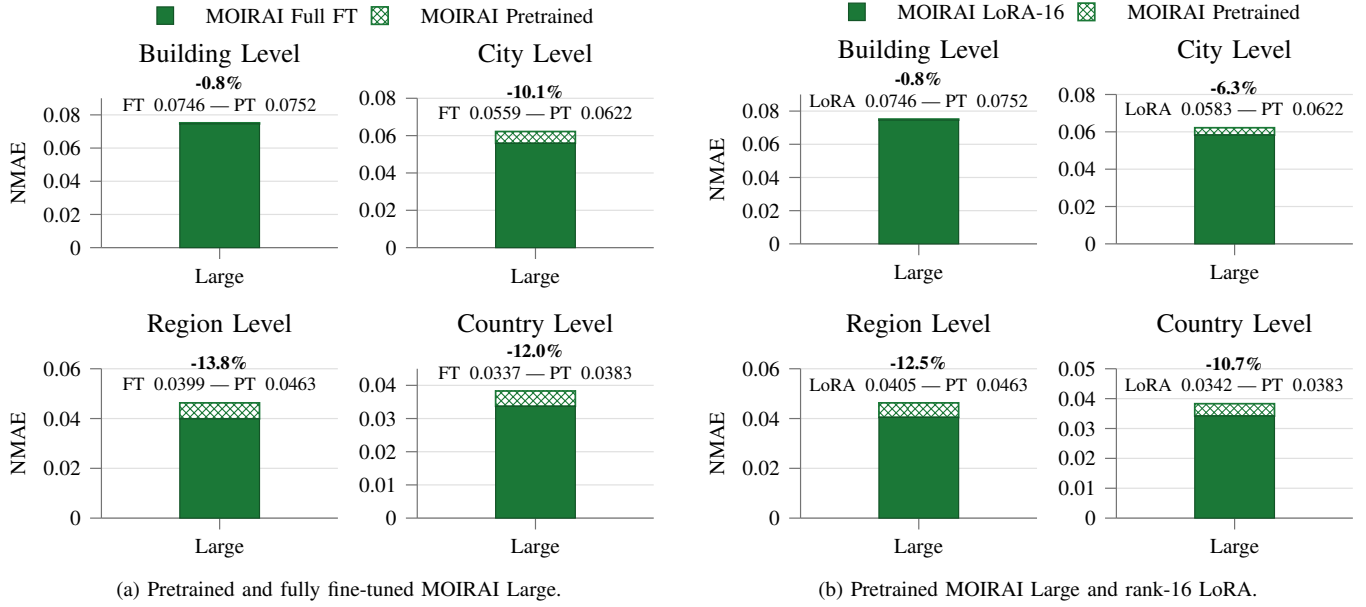
\begin{figure*}[t]
\centering
\begin{minipage}[t]{0.49\textwidth}
\centering
\centering

\definecolor{largemodel}{HTML}{1B7837}

\newcommand{\plotmoiraiimprovement}[3]{%
    \pgfmathsetmacro{\pretrainedsegment}{#2-#1}%

    \addplot+[
        ybar,
        fill=largemodel,
        draw=largemodel!70!black,
        line width=0.45pt
    ] coordinates {(Large,#1)};

    \addplot+[
        ybar,
        fill=white,
        draw=largemodel,
        line width=0.65pt,
        pattern=crosshatch,
        pattern color=largemodel
    ] coordinates {(Large,\pretrainedsegment)};

    \node[
        font=\scriptsize,
        text=black,
        anchor=south,
        yshift=1pt
    ] at (axis cs:Large,#2)
    {FT \,#1\;|\;PT \,#2};

    \node[
        font=\scriptsize\bfseries,
        text=black,
        anchor=south,
        yshift=9pt
    ] at (axis cs:Large,#2)
    {#3};
}

\begin{tikzpicture}
    \begin{groupplot}[
        group style={
            group size=2 by 2,
            horizontal sep=1.05cm,
            vertical sep=1.60cm,
        },
        width=0.53\linewidth,
        height=0.40\linewidth,
        ybar stacked,
        /pgf/bar width=30pt,
        /pgf/bar shift=0pt,
        ymin=0,
        scaled y ticks=false,
        symbolic x coords={Large},
        xtick={Large},
        xticklabels={Large},
        x tick label style={
            font=\footnotesize
        },
        yticklabel style={
            font=\footnotesize
        },
        tick align=outside,
        axis x line*=bottom,
        axis y line*=left,
        axis line style={
            draw=black!55,
            line width=0.45pt
        },
        tick style={
            draw=black!55,
            line width=0.45pt
        },
        ymajorgrids=true,
        grid style={
            draw=black!12,
            line width=0.45pt
        },
        xmin={[normalized]-0.5},
        xmax={[normalized]0.5},
        title style={
            font=\normalfont,
            yshift=2pt
        },
        ylabel style={
            font=\footnotesize
        },
        clip=false,
    ]

    \nextgroupplot[
        title={Building Level},
        ylabel={NMAE},
        ymax=0.09,
        ytick={0,0.02,0.04,0.06,0.08},
        yticklabels={0,0.02,0.04,0.06,0.08},
        legend to name=moiraiimprovementlegend,
        legend columns=2,
        legend style={
            draw=none,
            font=\footnotesize,
            column sep=0.35cm,
            /tikz/every even column/.append style={
                column sep=0.10cm
            },
            legend image code/.code={%
                \draw[##1]
                (0,-0.09cm)
                rectangle
                (0.22cm,0.14cm);%
            },
        },
    ]

    \plotmoiraiimprovement
        {0.0746}
        {0.0752}
        {-0.8\%}

    \addlegendentry{MOIRAI Full FT}
    \addlegendentry{MOIRAI Pretrained}

    \nextgroupplot[
        title={City Level},
        ymax=0.08,
        ytick={0,0.02,0.04,0.06,0.08},
        yticklabels={0,0.02,0.04,0.06,0.08},
    ]

    \plotmoiraiimprovement
        {0.0559}
        {0.0622}
        {-10.1\%}

    \nextgroupplot[
        title={Region Level},
        ylabel={NMAE},
        ymax=0.06,
        ytick={0,0.02,0.04,0.06},
        yticklabels={0,0.02,0.04,0.06},
    ]

    \plotmoiraiimprovement
        {0.0399}
        {0.0463}
        {-13.8\%}

    \nextgroupplot[
        title={Country Level},
        ymax=0.045,
        ytick={0,0.01,0.02,0.03,0.04},
        yticklabels={0,0.01,0.02,0.03,0.04},
    ]

    \plotmoiraiimprovement
        {0.0337}
        {0.0383}
        {-12.0\%}

    \end{groupplot}

    \node[anchor=south] at
    ($(group c1r1.north east)!0.5!
      (group c2r1.north west)+(0,0.68cm)$)
    {\pgfplotslegendfromname{moiraiimprovementlegend}};

\end{tikzpicture}
\vspace{0.2em}

\footnotesize (a) Pretrained and fully fine-tuned MOIRAI Large.
\end{minipage}\hfill
\begin{minipage}[t]{0.49\textwidth}
\centering
\centering

\definecolor{largemodel}{HTML}{1B7837}

\newcommand{\plotpeftimprovement}[3]{%
    \pgfmathsetmacro{\reductionsegment}{#2-#1}%

    \addplot+[
        ybar,
        fill=largemodel,
        draw=largemodel!70!black,
        line width=0.45pt
    ] coordinates {(Large,#1)};

    \addplot+[
        ybar,
        fill=white,
        draw=largemodel,
        line width=0.65pt,
        pattern=crosshatch,
        pattern color=largemodel
    ] coordinates {(Large,\reductionsegment)};

    \node[
        font=\scriptsize,
        text=black,
        anchor=south,
        yshift=1pt
    ] at (axis cs:Large,#2)
    {LoRA \,#1\;|\;PT \,#2};

    \node[
        font=\scriptsize\bfseries,
        text=black,
        anchor=south,
        yshift=9pt
    ] at (axis cs:Large,#2)
    {#3};
}

\begin{tikzpicture}
    \begin{groupplot}[
        group style={
            group size=2 by 2,
            horizontal sep=1.05cm,
            vertical sep=1.65cm,
        },
        width=0.53\linewidth,
        height=0.40\linewidth,
        ybar stacked,
        /pgf/bar width=30pt,
        /pgf/bar shift=0pt,
        ymin=0,
        scaled y ticks=false,
        symbolic x coords={Large},
        xtick={Large},
        xticklabels={Large},
        x tick label style={
            font=\footnotesize
        },
        yticklabel style={
            font=\footnotesize
        },
        tick align=outside,
        axis x line*=bottom,
        axis y line*=left,
        axis line style={
            draw=black!55,
            line width=0.45pt
        },
        tick style={
            draw=black!55,
            line width=0.45pt
        },
        ymajorgrids=true,
        grid style={
            draw=black!12,
            line width=0.45pt
        },
        xmin={[normalized]-0.5},
        xmax={[normalized]0.5},
        title style={
            font=\normalfont,
            yshift=2pt
        },
        ylabel style={
            font=\footnotesize
        },
        clip=false,
    ]

    \nextgroupplot[
        title={Building Level},
        ylabel={NMAE},
        ymax=0.09,
        ytick={0,0.02,0.04,0.06,0.08},
        yticklabels={0,0.02,0.04,0.06,0.08},
        legend to name=peftpretrainedlegend,
        legend columns=2,
        legend style={
            draw=none,
            font=\footnotesize,
            column sep=0.35cm,
            /tikz/every even column/.append style={
                column sep=0.10cm
            },
            legend image code/.code={%
                \draw[##1]
                (0,-0.09cm)
                rectangle
                (0.22cm,0.14cm);%
            },
        },
    ]

    \plotpeftimprovement
        {0.0746}
        {0.0752}
        {-0.8\%}

    \addlegendentry{MOIRAI LoRA-16}
    \addlegendentry{MOIRAI Pretrained}

    \nextgroupplot[
        title={City Level},
        ymax=0.08,
        ytick={0,0.02,0.04,0.06,0.08},
        yticklabels={0,0.02,0.04,0.06,0.08},
    ]

    \plotpeftimprovement
        {0.0583}
        {0.0622}
        {-6.3\%}

    \nextgroupplot[
        title={Region Level},
        ylabel={NMAE},
        ymax=0.06,
        ytick={0,0.02,0.04,0.06},
        yticklabels={0,0.02,0.04,0.06},
    ]

    \plotpeftimprovement
        {0.0405}
        {0.0463}
        {-12.5\%}

    \nextgroupplot[
        title={Country Level},
        ymax=0.05,
        ytick={0,0.01,0.02,0.03,0.04,0.05},
        yticklabels={0,0.01,0.02,0.03,0.04,0.05},
    ]

    \plotpeftimprovement
        {0.0342}
        {0.0383}
        {-10.7\%}

    \end{groupplot}

    \node[anchor=south] at
    ($(group c1r1.north east)!0.5!
      (group c2r1.north west)+(0,0.68cm)$)
    {\pgfplotslegendfromname{peftpretrainedlegend}};

\end{tikzpicture}
\vspace{0.2em}

\footnotesize (b) Pretrained MOIRAI Large and rank-16 \gls{lora}.
\end{minipage}
\caption{Within-model effect of full fine-tuning and rank-16 \gls{lora} across electrical load aggregation levels.}
\label{fig:moirai_adaptation}
\end{figure*}

Fig.~\ref{fig:moirai_adaptation} separates the effects of full fine-tuning and rank-16 \gls{lora} so that the adaptation gains remain visible within the MOIRAI family. Relative to pretrained MOIRAI Large, full fine-tuning reduces \gls{nmae} by approximately 0.8\,\%, 10.1\,\%, 13.8\,\%, and 12.0\,\% for building-level, city-level, regional-level, and country-level data. Rank-16 \gls{lora} follows the same pattern, with reductions of approximately 0.8\,\%, 6.3\,\%, 12.5\,\%, and 10.7\,\%. Averaged across the four levels, \gls{lora} improves pretrained MOIRAI by approximately 7.5\,\%. Across the four levels, rank-16 \gls{lora} retains most of the improvement achieved by full fine-tuning, although full fine-tuning provides a clearer advantage at city level.

Both full fine-tuning and \gls{lora} produce only limited improvement at the building level, suggesting that adapter capacity is not the only limiting factor. Historical load and temperature alone may not fully represent the heterogeneous functions, schedules, occupancy patterns, and equipment states that influence building demand. For these targets, broader building coverage and more informative operational covariates may be more valuable than updating additional model parameters.

Rank-16 \gls{lora} updates 3.146 million parameters, approximately 1\,\% of MOIRAI Large, compared with all 311 million parameters under full fine-tuning. The \gls{lora} and full fine-tuning experiments were executed on an NVIDIA V100 with 16~GB and an NVIDIA A100 with 80~GB, respectively. These values describe the hardware used rather than the minimum memory required by either method.

The broader ablation results support rank-16 \gls{lora} as a practical default for MOIRAI. Increasing the rank from 16 to 128 raises the number of trainable parameters from 3.146 million to 25.166 million without producing a consistent improvement across aggregation levels. \gls{lora}, \gls{qlora}, and \gls{dora} achieve similar mean \gls{nmae}, with no method showing a consistent advantage. A broader building-level training set improves building-level accuracy but has mixed effects on the other aggregation levels. These findings shift the practical focus from adapter complexity toward pretrained model strength, representative training data, and validation on the intended electrical load targets.

\subsection{Cross-Model In-Distribution Comparison}
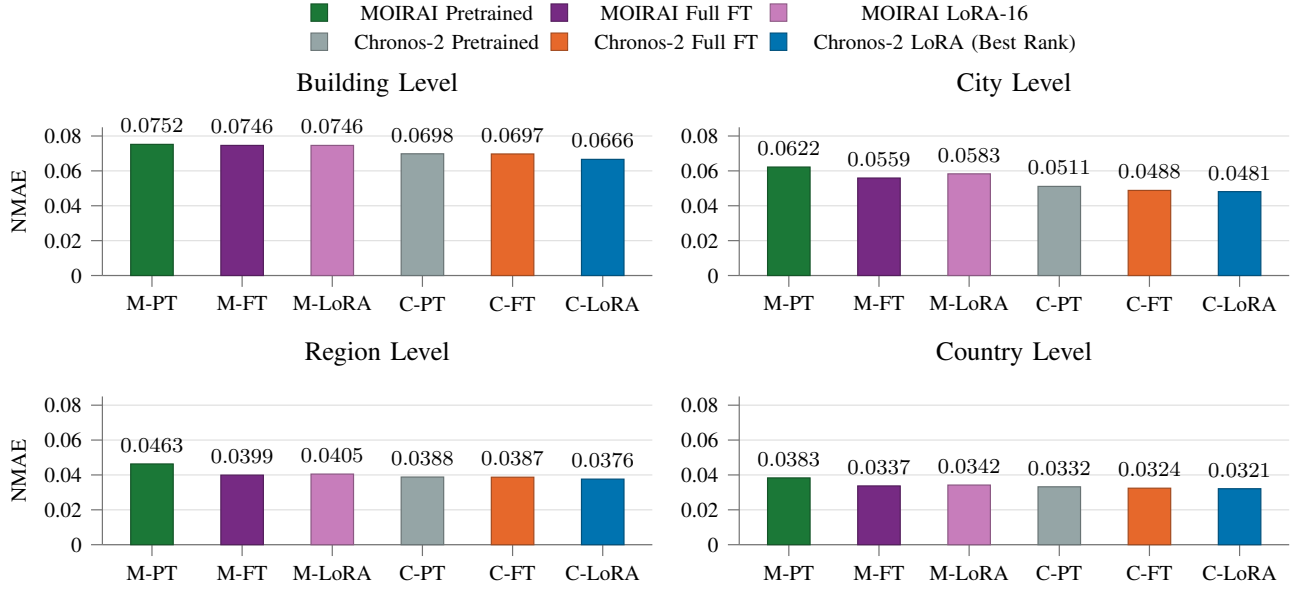
\begin{figure*}[t]
\centering
\centering

\definecolor{moiraipt}{HTML}{1B7837}
\definecolor{moiraift}{HTML}{762A83}
\definecolor{moirailora}{HTML}{C77DBB}
\definecolor{chronospt}{HTML}{95A5A6}
\definecolor{chronosft}{HTML}{E6682B}
\definecolor{chronoslora}{HTML}{0073B0}

\newcommand{\plotcomparisonbar}[3]{%
    \addplot+[
        fill=#3,
        draw=#3!70!black,
        line width=0.45pt
    ] coordinates {(#1,#2)};
}

\begin{tikzpicture}
    \begin{groupplot}[
        group style={
            group size=2 by 2,
            horizontal sep=1.15cm,
            vertical sep=1.6cm,
        },
        width=\columnwidth,
        height=0.40\columnwidth,
        ybar=0pt,
        /pgf/bar width=16pt,
        /pgf/bar shift=0pt,
        ymin=0,
        ymax=0.085,
        ytick={0,0.02,0.04,0.06,0.08},
        yticklabels={0,0.02,0.04,0.06,0.08},
        scaled y ticks=false,
        symbolic x coords={
            M-PT,
            M-FT,
            M-LoRA,
            C-PT,
            C-FT,
            C-LoRA
        },
        xtick={
            M-PT,
            M-FT,
            M-LoRA,
            C-PT,
            C-FT,
            C-LoRA
        },
        xticklabels={
            M-PT,
            M-FT,
            M-LoRA,
            C-PT,
            C-FT,
            C-LoRA
        },
        x tick label style={
            font=\footnotesize
        },
        yticklabel style={
            font=\footnotesize
        },
        tick align=outside,
        axis x line*=bottom,
        axis y line*=left,
        axis line style={
            draw=black!55,
            line width=0.45pt
        },
        tick style={
            draw=black!55,
            line width=0.45pt
        },
        ymajorgrids=true,
        grid style={
            draw=black!12,
            line width=0.45pt
        },
        enlarge x limits=0.11,
        nodes near coords={%
            \pgfmathprintnumber[
                fixed,
                precision=4,
                fixed zerofill
            ]{\pgfplotspointmeta}%
        },
        nodes near coords style={
            font=\footnotesize,
            text=black,
            yshift=1pt
        },
        title style={
            font=\normalfont,
            yshift=2pt
        },
        ylabel style={
            font=\footnotesize
        },
        clip=false,
    ]

    \nextgroupplot[
        title={Building Level},
        ylabel={NMAE},
        legend to name=overallcomparisonlegend,
        legend columns=3,
        legend style={
            draw=none,
            font=\footnotesize,
            column sep=0.28cm,
            /tikz/every even column/.append style={
                column sep=0.08cm
            },
            legend image code/.code={%
                \draw[#1]
                (0,-0.09cm) rectangle (0.22cm,0.14cm);%
            },
        },
    ]

    \addlegendimage{fill=moiraipt, draw=moiraipt!70!black}
    \addlegendentry{MOIRAI Pretrained}
    \addlegendimage{fill=moiraift, draw=moiraift!70!black}
    \addlegendentry{MOIRAI Full FT}
    \addlegendimage{fill=moirailora, draw=moirailora!70!black}
    \addlegendentry{MOIRAI LoRA-16}
    \addlegendimage{fill=chronospt, draw=chronospt!70!black}
    \addlegendentry{Chronos-2 Pretrained}
    \addlegendimage{fill=chronosft, draw=chronosft!70!black}
    \addlegendentry{Chronos-2 Full FT}
    \addlegendimage{fill=chronoslora, draw=chronoslora!70!black}
    \addlegendentry{Chronos-2 LoRA (Best Rank)}

    \plotcomparisonbar{M-PT}{0.0752}{moiraipt}
    \plotcomparisonbar{M-FT}{0.0746}{moiraift}
    \plotcomparisonbar{M-LoRA}{0.0746}{moirailora}
    \plotcomparisonbar{C-PT}{0.0698}{chronospt}
    \plotcomparisonbar{C-FT}{0.0697}{chronosft}
    \plotcomparisonbar{C-LoRA}{0.0666}{chronoslora}

    \nextgroupplot[
        title={City Level},
    ]

    \plotcomparisonbar{M-PT}{0.0622}{moiraipt}
    \plotcomparisonbar{M-FT}{0.0559}{moiraift}
    \plotcomparisonbar{M-LoRA}{0.0583}{moirailora}
    \plotcomparisonbar{C-PT}{0.0511}{chronospt}
    \plotcomparisonbar{C-FT}{0.0488}{chronosft}
    \plotcomparisonbar{C-LoRA}{0.0481}{chronoslora}

    \nextgroupplot[
        title={Region Level},
        ylabel={NMAE},
    ]

    \plotcomparisonbar{M-PT}{0.0463}{moiraipt}
    \plotcomparisonbar{M-FT}{0.0399}{moiraift}
    \plotcomparisonbar{M-LoRA}{0.0405}{moirailora}
    \plotcomparisonbar{C-PT}{0.0388}{chronospt}
    \plotcomparisonbar{C-FT}{0.0387}{chronosft}
    \plotcomparisonbar{C-LoRA}{0.0376}{chronoslora}

    \nextgroupplot[
        title={Country Level},
    ]

    \plotcomparisonbar{M-PT}{0.0383}{moiraipt}
    \plotcomparisonbar{M-FT}{0.0337}{moiraift}
    \plotcomparisonbar{M-LoRA}{0.0342}{moirailora}
    \plotcomparisonbar{C-PT}{0.0332}{chronospt}
    \plotcomparisonbar{C-FT}{0.0324}{chronosft}
    \plotcomparisonbar{C-LoRA}{0.0321}{chronoslora}

    \end{groupplot}

    \node[anchor=south] at
    ($(group c1r1.north east)!0.5!(group c2r1.north west)+(0,0.72cm)$)
    {\pgfplotslegendfromname{overallcomparisonlegend}};

\end{tikzpicture}
\caption{Final \gls{id} \gls{nmae} comparison of pretrained, fully fine-tuned, and \gls{lora}-adapted MOIRAI 1.0 Large and Chronos-2 across the four electrical load aggregation levels. MOIRAI uses rank 16, while the Chronos-2 \gls{lora} result at each level denotes the lowest test \gls{nmae} observed in the retrospective rank comparison. All panels use the same scale to support direct comparison.}
\label{fig:final_id}
\end{figure*}

Fig.~\ref{fig:final_id} compares the selected MOIRAI Large and Chronos-2 configurations on the same \gls{nmae} scale because adaptation gains depend on the pretrained starting point. Chronos-2 already outperforms MOIRAI Large before adaptation, leaving less room for downstream improvement.

Relative to pretrained Chronos-2, full fine-tuning reduces \gls{nmae} by approximately 0.1\,\%, 4.5\,\%, 0.3\,\%, and 2.4\,\% for building-level, city-level, regional-level, and country-level data, respectively. \gls{lora} reduces \gls{nmae} by approximately 4.6\,\%, 5.9\,\%, 3.1\,\%, and 3.3\,\% when using the best-performing tested rank at each level. Rank 128 gives the lowest \gls{nmae} for building-level and regional-level data, while rank 64 gives the lowest \gls{nmae} for city-level and country-level data. Averaged across the four aggregation levels, full fine-tuning and \gls{lora} reduce \gls{nmae} by approximately 1.8\,\% and 4.2\,\%, respectively. In this retrospective comparison, the lowest observed \gls{lora} test \gls{nmae} is slightly below that of full fine-tuning at all four levels, although the absolute differences remain small. Across both backbones, pretrained error falls with aggregation, but adaptation gains differ: MOIRAI benefits mainly at city, regional, and country levels, whereas Chronos-2 gains most at building and city levels.

\subsection{Out-of-Distribution Transfer}
Table~\ref{tab:ood_metrics} compares \gls{nmae} and \gls{nrmse} across eight \gls{ood} series, each spanning one year, using a fixed 2048-hour context, a 96-hour forecast horizon, and a 96-hour stride. For Chronos-2, using the lowest observed \gls{ood} error across the four evaluated \gls{lora} ranks for each target and metric reduces mean \gls{nmae} from 0.0636 to 0.0504 and mean \gls{nrmse} from 0.1205 to 0.0960. Ci-AVA shows the largest improvement, with \gls{nmae} decreasing from 0.0459 to 0.0207 and \gls{nrmse} from 0.0732 to 0.0332. Both metrics also improve for BDS3--BDS5, whereas the pretrained checkpoint remains strongest for BDS1 and BDS2. C-IRE and R-DNK change only slightly. For MOIRAI, rank 16 already gives the lowest \gls{nmae} for five of the eight targets, while higher ranks provide only marginal improvements for the remaining three; the lowest-error Chronos-2 rank varies more across targets. The agreement between the two metrics indicates that adaptation benefits vary by target, with the clearest gains for the local-area series and selected buildings.

\begin{table*}[!ht]
\caption{Out-of-distribution NMAE and NRMSE using the fixed 2048-hour context. Lower values indicate better performance. Bold and underlined values denote the best and second-best results, respectively.}
\label{tab:ood_metrics}
\centering
\footnotesize
\setlength{\tabcolsep}{7.0pt}
\renewcommand{\arraystretch}{1.00}
\begin{tabular}{llcccccccc}
\toprule
Method & Metric & C-IRE & R-DNK & Ci-AVA & BDS1 & BDS2 & BDS3 & BDS4 & BDS5 \\
\midrule

\multirow{2}{*}{Seasonal Naive}
& NMAE  & 0.0302 & 0.0584 & 0.0672 & 0.0768 & 0.0955 & 0.1037 & 0.1672 & 0.1424 \\
& NRMSE & 0.0460 & 0.0902 & 0.1024 & 0.1280 & 0.1742 & 0.2158 & 0.3670 & 0.2484 \\
\midrule

\multirow{2}{*}{XGBoost}
& NMAE  & 0.0371 & 0.0713 & 0.0806 & 0.0749 & 0.0868 & 0.1121 & 0.1676 & 0.1197 \\
& NRMSE & 0.0562 & 0.1091 & 0.1274 & 0.1221 & 0.1560 & 0.2267 & 0.3769 & 0.2129 \\
\midrule

\multirow{2}{*}{MOIRAI Pretrained}
& NMAE  & 0.0246 & 0.0456 & 0.0503 & 0.0573 & 0.0782 & 0.0792 & 0.1115 & 0.0984 \\
& NRMSE & 0.0375 & 0.0709 & 0.0783 & 0.0972 & 0.1403 & 0.1650 & 0.2457 & 0.1755 \\
\midrule

\multirow{2}{*}{MOIRAI LoRA}
& NMAE  & 0.0224 & 0.0435 & 0.0468 & \underline{0.0520} & 0.0718 & 0.0768 & 0.1110 & 0.0960 \\
& NRMSE & 0.0346 & 0.0676 & 0.0729 & \underline{0.0882} & 0.1306 & 0.1611 & 0.2469 & 0.1707 \\
\midrule

\multirow{2}{*}{MOIRAI Full FT}
& NMAE  & 0.0227 & 0.0437 & 0.0467 & 0.0585 & 0.0755 & 0.0787 & 0.1150 & 0.0965 \\
& NRMSE & 0.0349 & 0.0680 & 0.0732 & 0.0983 & 0.1368 & 0.1651 & 0.2585 & 0.1709 \\
\midrule

\multirow{2}{*}{Chronos-2 Pretrained}
& NMAE  & 0.0210 & 0.0415 & 0.0459 & \textbf{0.0494} & \textbf{0.0660} & 0.0768 & 0.1112 & 0.0972 \\
& NRMSE & 0.0330 & 0.0645 & 0.0732 & \textbf{0.0848} & \textbf{0.1236} & 0.1627 & 0.2483 & 0.1740 \\
\midrule

\multirow{2}{*}{Chronos-2 LoRA}
& NMAE  & \textbf{0.0200} & \underline{0.0405} & \textbf{0.0207} & 0.0530 & \underline{0.0665} & \underline{0.0603} & \textbf{0.0688} & \textbf{0.0731} \\
& NRMSE & \textbf{0.0315} & \underline{0.0630} & \textbf{0.0332} & 0.0907 & \underline{0.1239} & \underline{0.1293} & \textbf{0.1647} & \textbf{0.1316} \\
\midrule

\multirow{2}{*}{Chronos-2 Full FT}
& NMAE  & \underline{0.0204} & \textbf{0.0404} & \underline{0.0218} & 0.0521 & 0.0679 & \textbf{0.0601} & \underline{0.0698} & \underline{0.0737} \\
& NRMSE & \underline{0.0319} & \textbf{0.0627} & \underline{0.0352} & 0.0895 & 0.1256 & \textbf{0.1288} & \underline{0.1678} & \underline{0.1326} \\
\bottomrule
\end{tabular}

\vspace{1.0ex}
\begin{minipage}{0.98\textwidth}
\footnotesize
C-IRE, R-DNK, and Ci-AVA denote the national load of Ireland, the regional load of Denmark, and the AVA local-area series, respectively; BDS1--BDS5 denote withheld Building Data Genome 2 buildings. All four \gls{lora} ranks were evaluated on the \gls{ood} series. For both MOIRAI and Chronos-2, \gls{lora} reports the lowest observed \gls{ood} error across ranks for each target and metric.
\end{minipage}

\end{table*}

\begin{figure*}[!t]
\centering
\input{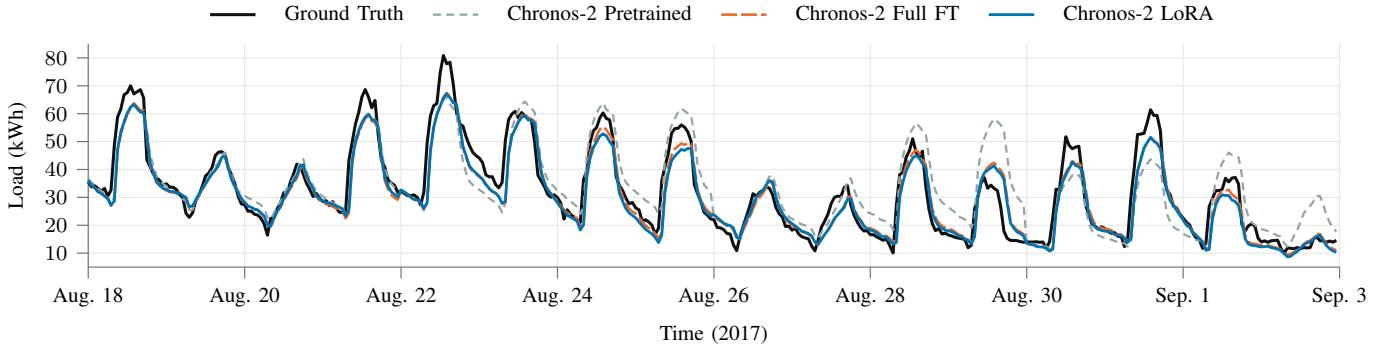}
\vspace{-5mm}
\caption{Illustrative contiguous \gls{ood} test segment from a withheld healthcare building, containing rapid load transitions and peak-demand behavior. The fully fine-tuned and \gls{lora}-adapted Chronos-2 variants follow the observed amplitude changes more closely than the pretrained checkpoint.}
\label{fig:bds4_line}
\end{figure*}

Fig.~\ref{fig:bds4_line} illustrates one building-level case using a withheld healthcare building. Pretrained Chronos-2 captures the broad daily pattern but smooths rapid transitions and underestimates several peaks. The adapted variants follow changes in demand amplitude more closely. The \gls{nrmse} results provide complementary evidence that larger forecast errors decrease for BDS3--BDS5, although BDS1 and BDS2 do not improve. The building-level benefit therefore remains target-dependent.

The \gls{id} and \gls{ood} findings do not identify one universally best adaptation configuration. \gls{id} improvement demonstrates specialization to the load sources represented during adaptation, whereas performance on a new target depends on whether the learned electrical load structure transfers to its geography, consumer mix, or building function. Operators should therefore validate an adapted checkpoint on representative external data before deploying it as the default for a new system.

\subsection{Practical Adaptation Guide}
\begin{enumerate}
    \item Evaluate the strongest pretrained checkpoint and retain it when performance is sufficient.
    \item Start with a compact \gls{lora} configuration and retain the adapted model only when validation shows a meaningful gain, then verify performance on external systems or buildings.
    \item Consider larger ranks or full fine-tuning only when validation shows a meaningful additional accuracy gain.
\end{enumerate}
For irregular building profiles, the deployment process should first consider broader training coverage and informative covariates before increasing adapter complexity, as recent building-energy results show that covariate integration can provide substantial gains without additional fine-tuning \cite{kreusel2026covariates}.

\section{Limitations}
The selected datasets, 96-hour forecast horizon, fixed 2048-hour context, model families, and adaptation settings limit the generalizability of the findings. The large adaptation experiments were not repeated across seeds, and the \gls{id} and \gls{ood} rank comparisons are retrospective test-set ablations.

\gls{nrmse} complements \gls{nmae} by giving greater weight to larger forecast errors, but neither metric directly quantifies forecast bias, peak timing, or probabilistic calibration. The city-level and building-level experiments assume access to temperature over the complete forecast horizon. The source datasets do not consistently provide occupancy, equipment, tariff, holiday, or operating-schedule information.

\section{Conclusion}
This study examined when parameter-efficient adaptation is worthwhile for electrical load forecasting. Pretrained \glspl{tsfm} outperform the implemented Seasonal Naive and XGBoost baselines across the evaluated aggregation levels. On MOIRAI Large, rank-16 \gls{lora} retains most of the full fine-tuning improvement while updating approximately 1\,\% of the model. For Chronos-2, the lowest observed \gls{lora} \gls{id} test \gls{nmae} is slightly lower than that of full fine-tuning across all four levels, although the absolute differences remain small. Its stronger pretrained performance also leaves less room for downstream improvement than MOIRAI.

The \gls{ood} \gls{nmae} and \gls{nrmse} results show that adaptation benefits vary by target, with the clearest adaptation gains for the local-area series and several buildings.

Adaptation is therefore not automatically necessary. Pretrained inference can perform strongly for smoother country-level and regional-level demand, whereas adaptation gains for local-area and building targets are more target-dependent and vary with the pretrained backbone. The resulting decision rule is to start with the strongest pretrained checkpoint and introduce compact adaptation only when representative validation shows a clear target-specific gain; the adapted checkpoint should then be verified on external systems or buildings before deployment.

\section*{Acknowledgment}
\noindent
This project was funded in part by the Deutsche Forschungsgemeinschaft (DFG, German Research Foundation) - 535389056.

\balance
\bibliographystyle{IEEEtran} 
\bibliography{literature}

@inproceedings{woo2024moirai,
  author    = {Woo, Gerald and Liu, Chenghao and Kumar, Akshat and Xiong, Caiming and Savarese, Silvio and Sahoo, Doyen},
  title     = {Unified Training of Universal Time Series Forecasting Transformers},
  booktitle = {Proceedings of the 41st International Conference on Machine Learning},
  volume    = {235},
  pages     = {53140--53164},
  year      = {2024}
}

@article{ansari2024chronos,
  author  = {Ansari, Abdul Fatir and others},
  title   = {{Chronos}: Learning the Language of Time Series},
  journal = {Transactions on Machine Learning Research},
  year    = {2024}
}

@misc{ansari2025chronos2,
  author = {Ansari, Abdul Fatir and others},
  title  = {{Chronos-2}: From Univariate to Universal Forecasting},
  year   = {2025},
  note   = {arXiv:2510.15821}
}

@inproceedings{hu2022lora,
  author    = {Hu, Edward J. and others},
  title     = {{LoRA}: Low-Rank Adaptation of Large Language Models},
  booktitle = {International Conference on Learning Representations},
  year      = {2022}
}

@article{sevlian2018,
  author  = {Sevlian, Raffi and Rajagopal, Ram},
  title   = {A Scaling Law for Short Term Load Forecasting on Varying Levels of Aggregation},
  journal = {International Journal of Electrical Power \& Energy Systems},
  volume  = {98},
  pages   = {350--361},
  year    = {2018},
  doi = {10.1016/j.ijepes.2017.10.032},
}

@inproceedings{peng2019,
  author    = {Peng, Yayu and Wang, Yishen and Lu, Xiao and Li, Haifeng and Shi, Di and Wang, Zhiwei   and Li, Jie},
  title     = {Short-Term Load Forecasting at Different Aggregation Levels with Predictability     Analysis},
  doi = {10.1109/ISGT-Asia.2019.8881343},
  booktitle = {2019 IEEE Innovative Smart Grid Technologies--Asia ({ISGT Asia})},
  pages     = {3385--3390},
  year      = {2019}
}

@inproceedings{chen2016xgboost,
  author    = {Chen, Tianqi and Guestrin, Carlos},
  title     = {{XGBoost}: A Scalable Tree Boosting System},
  booktitle = {Proceedings of the 22nd ACM SIGKDD International Conference on Knowledge Discovery and Data Mining},
  doi = {10.1145/2939672.2939785},
  pages     = {785--794},
  year      = {2016}
}

@inproceedings{gupta2024lora,
  author    = {Gupta, D. and others},
  title     = {Low-Rank Adaptation of Time Series Foundational Models for Out-of-Domain Modality Forecasting},
  booktitle = {Proceedings of the 26th International Conference on Multimodal Interaction},
  doi = {10.1145/3678957.3685724},
  pages     = {382--386},
  year      = {2024}
}

@article{park2025building,
  author  = {Park, Young-Jin and others},
  title   = {Probabilistic Forecasting for Building Energy Systems
 Using Time-Series Foundation Models},
  journal = {Energy and Buildings},
  doi = {10.1016/j.enbuild.2025.116446},
  volume  = {348},
  pages   = {116446},
  year    = {2025}
}

@misc{ullah2026weld,
  author       = {Ullah, Md Habib and others},
  title        = {Worldwide Electricity Load Dataset},
  howpublished = {Mendeley Data, Version 1},
  year         = {2026},
  doi          = {10.17632/ybggkc58fz.1}
}

@misc{salam2018tetouan,
  author       = {Salam, Abdulwahed and El Hibaoui, Abdelaaziz},
  title        = {Power Consumption of Tetouan City},
  howpublished = {UCI Machine Learning Repository},
  year         = {2018},
  doi          = {10.24432/C5B034}
}

@article{miller2020bdg2,
  author  = {Miller, Clayton and others},
  title   = {The Building Data Genome Project 2, Energy Meter Data from the {ASHRAE} Great Energy  Predictor {III} Competition},
  journal = {Scientific Data},
  doi = {10.1038/s41597-020-00712-x},
  volume  = {7},
  pages   = {368},
  year    = {2020}
}

@inproceedings{dettmers2023qlora,
  author    = {Dettmers, Tim and Pagnoni, Artidoro and Holtzman, Ari and Zettlemoyer, Luke},
  title     = {{QLoRA}: Efficient Finetuning of Quantized {LLMs}},
  booktitle = {Advances in Neural Information Processing Systems},
  volume    = {36},
  year      = {2023},
  pages = {10088--10115},
  doi   = {10.52202/075280-0441},
}

@inproceedings{liu2024dora,
  author    = {Liu, Shih-Yang and others},
  title     = {{DoRA}: Weight-Decomposed Low-Rank Adaptation},
  booktitle = {Proceedings of the 41st International Conference on Machine Learning},
  volume    = {235},
  pages     = {32100--32121},
  year      = {2024}
}

@inproceedings{chattopadhyay2026peak,
  author    = {Chattopadhyay, Souhardya and Oelhaf, Julian and Schoening, Antonia and Deuschel, Jessica and Bhattacharyya, Bitan and Bergler, Christian and Maier, Andreas and Bayer, Siming},
  title     = {Peak-Aware Short-Term Load Forecasting Across Distribution Grid Aggregation Levels},
  booktitle = {2026 IEEE PES Innovative Smart Grid Technologies Conference Europe (ISGT Europe)},
  year      = {2026},
  month     = oct,
  address   = {Budapest, Hungary},
  publisher = {IEEE},
  note      = {arXiv:2609.18588}, 
}

@inproceedings{arango2025chronosx,
    title =  {{ChronosX}: Adapting Pretrained Time Series Models with Exogenous Variables},
    author = {Arango, Sebastian Pineda and others},
    booktitle =  {Proceedings of The 28th International Conference on Artificial Intelligence and Statistics},
    pages =  {2242--2250},
    year =  {2025},
    editor =  {Li, Yingzhen and Mandt, Stephan and Agrawal, Shipra and Khan, Emtiyaz},  volume =  {258},
    series =  {Proceedings of Machine Learning Research},
    month =  {03--05 May},
    publisher =    {PMLR},  url =  {https://proceedings.mlr.press/v258/arango25a.html},
}

@inproceedings{kreusel2026covariates,
  author    = {Kreusel, Alexander and Hertel, Matthias and Noskiewicz, Moritz and Zahn, Frederik and Maa{\ss}, Heiko and Mikut, Ralf and Hagenmeyer, Veit},
  title     = {Covariates Are the Key to Accurate Probabilistic Building Energy
               Forecasting with Time Series Foundation Models},
  booktitle = {Proceedings of the 2026 ACM Sustainability Week},
  pages     = {172--182},
  year      = {2026},
  publisher = {Association for Computing Machinery},
  doi       = {10.1145/3765611.3815507}
}

\end{document}